\documentclass[letterpaper, 10pt, conference]{ieeeconf}  

\IEEEoverridecommandlockouts
\usepackage{xparse} 
\usepackage{amsmath} 
\usepackage{amssymb}
\usepackage{amsfonts} 
\usepackage{bbm}

\usepackage{graphicx}

\usepackage{hyperref}
\usepackage{cleveref}

\usepackage[backend=bibtex,style=ieee,natbib=true]{biblatex} 
\usepackage{nomencl} 
\makenomenclature

\NewDocumentCommand\placeholder{mmo}{
  \framebox{
    \begin{minipage}[c][#2]{#1}
      \centering
      \IfNoValueTF{#3}{Placeholder}{#3}
    \end{minipage} 
  }
}

\usepackage{graphicx}
\usepackage{amsmath,amssymb} 
\usepackage{color}
\usepackage{xparse} 
\usepackage{amsmath} 
\usepackage{amssymb}
\usepackage{amsfonts} 
\usepackage{bbm}
\usepackage{graphicx}
\usepackage{caption}
\usepackage{subcaption}
\usepackage{tabularx}
\usepackage{booktabs}
\usepackage{multirow}
\usepackage{textcomp}
\usepackage[font=small]{caption}
\usepackage{algorithm} 
\usepackage[noend]{algpseudocode}

\usepackage{tikz}
\usepackage[customcolors]{hf-tikz} 
\usetikzlibrary{patterns}
\usetikzlibrary{matrix,decorations.pathreplacing}
\pgfkeys{tikz/mymatrixenv/.style={decoration={brace},every left delimiter/.style={xshift=8pt},every right delimiter/.style={xshift=-8pt}}}
\pgfkeys{tikz/mymatrix/.style={matrix of math nodes,nodes in empty cells,left delimiter={[},right delimiter={]},inner sep=1pt,outer sep=1.5pt,column sep=2pt,row sep=2pt,nodes={minimum width=20pt,minimum height=10pt,anchor=center,inner sep=0pt,outer sep=0pt}}}
\pgfkeys{tikz/mymatrixbrace/.style={decorate,thick}}

\title{\LARGE \bf Online Descriptor Enhancement via Self-Labelling Triplets for Visual Data Association}

\author{Yorai Shaoul, Katherine Liu, Kyel Ok, and Nicholas Roy$^{}$
\thanks{$^{}$All authors are with the Computer Science and Artificial Intelligence Laboratory, Massachusetts Institute of Technology in Cambridge, USA.
{\tt\small \{yorai, katliu,kyelok,nickroy\}@mit.edu}}%
\thanks{$^{}$This research was sponsored by the MIT Quest for Intelligence and the Army Research
Laboratory. It was accomplished under
Cooperative Agreement Number W911NF-17-2-0181.  Their support is
gratefully acknowledged. }%
}

\DeclareMathOperator*{\argmin}{arg\,min}
\newcommand{\desc}{\mathbf{d}}
\newcommand{\descgen}{\mathcal{F}}
\newcommand{\patch}{p}

\newcommand{\varSizeBatch}{N}

\begin{document}

\maketitle
\thispagestyle{empty}
\pagestyle{empty}

\begin{abstract}

\textcolor{black}{Object-level data association is central to robotic applications such as tracking-by-detection and object-level simultaneous localization and mapping. While current learned visual data association methods outperform hand-crafted algorithms, many rely on large collections of domain-specific training examples that can be difficult to obtain without prior knowledge. Additionally, such methods often remain fixed during inference-time and do not harness observed information to better their performance.}
We propose a self-supervised method for incrementally refining visual descriptors to improve performance in the task of object-level visual data association. Our method optimizes deep descriptor generators online, \textcolor{black}{by continuously training a widely available image classification network pre-trained with domain-independent data.} We show that earlier layers in the network outperform later-stage layers for the data association task while also allowing for a 94\% reduction in the number of parameters, enabling the online optimization. We show that \textcolor{black}{self-labelling challenging triplets--}choosing positive examples separated by large temporal distances and negative examples close in the descriptor space\textcolor{black}{--}improves the quality of the learned descriptors for the multi-object tracking task. 
Finally, we demonstrate \textcolor{black}{that our approach surpasses other visual data-association methods applied to a tracking-by-detection task, and show that it provides better performance-gains when compared to other methods that attempt to adapt to observed information.}


\end{abstract}

\section{Introduction}
We are interested in matching visual object detections across temporally separated frames -- a fundamental capability for a wide range of applications in robotics and computer vision such as object tracking-by-detection and object-level simultaneous localization and mapping \cite{thrun2008simultaneous, ok2019robust}.

Although supervised learning methods \cite{holliday2017long, zagoruyko2015learning} have recently outperformed hand-engineered descriptors when attempting to adapt robustly to new data-association problems \cite{lowe2004sift, bay2006surf}, they can be difficult to train, and perform inconsistently. Relying on the existence of massive labeled datasets containing domain-specific training samples, supervised learning of descriptors or affinity metrics may fall short when deployed to novel environments \cite{fischer2014descriptor}. Contradictory results in the literature also shed light on the inconsistencies plaguing descriptors generated with pre-trained models \cite{balntas2017hpatches}.

In contrast, self-supervised learning methods aim to reduce \cite{ma2018customized} or eliminate annotation requirements and improve solutions online \cite{fischer2014descriptor}. To this end, these methods harness the temporal structure of video sequences to collect and annotate positive pairs of image-patches (i.e., subsets of frame pixels) containing the same object in real time, and compile those with negative samples (patches of different objects) into training datasets. The reliance solely on visual information removes the need for relevant annotated data, which may be difficult to obtain for novel environments.

\begin{figure}[tbp]
\centerline{\includegraphics[scale=0.54]{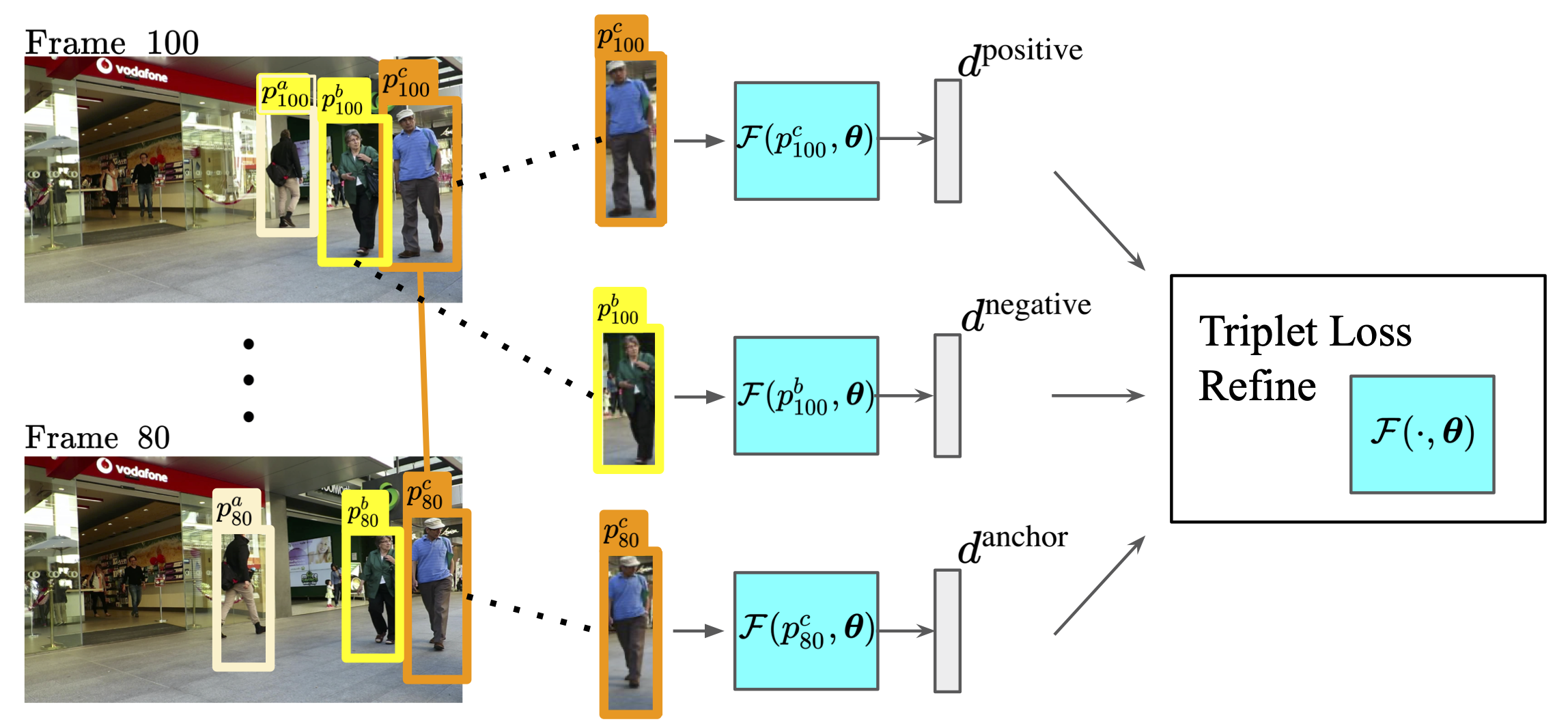}}
\caption{Our proposed approach self-supervises label generation to incrementally optimize a deep descriptor generator (cyan). To construct a triplet when frame 100 is received, we choose a positive (i.e., correct) object detection ($p^c_{100}$), a temporally distant anchor instance of the same object ($p^c_{80}$), and a negative example from the same frame that is closest in the current descriptor space ($p^b_{100}$). When enough patch-triplets are aggregated, they are used to train a descriptor-generator as a batch.
The visuals are frames 80 to 100 of the sequence ADL-Rundle-6 included in \cite{leal2015motchallenge}.}
\label{fig_triplet_gen}
\vspace{-0.3in}
\end{figure}

However, due to the computational overhead of online training, one of the challenges of online self-supervised learning for data association is generating training data which are compact but informative. Although the importance of finding informative negative samples is generally acknowledged, many existing approaches rely on simple heuristics such as randomly sampling training examples from candidates \cite{yoon2018online, bae2014robust, ma2018customized}. Other approaches consider image-space properties such as bounding box overlaps \cite{bewley2016alextrac, ma2018customized}.

In addition, existing approaches tend to still be reliant on pre-trained models, which can limit performance if assumed to be static, or be difficult to obtain if they require offline training. For example, many approaches for tractable online self-supervised visual data association methods have focused on learning lightweight affinity metrics (e.g. via logistic regression) between pairs of patch descriptors \cite{bewley2016alextrac, weinberger2009distance, solera2015learning, yoon2018online}. Without a mechanism for updating the descriptor generation model, such approaches are upper bounded by the representational power of their pre-trained models, and may struggle to extend to novel scenarios. Other methods learn descriptors online, but require labelled detection pairs to pre-train custom descriptor networks before performing online refinement, which may be difficult to obtain for arbitrary new environments \cite{bae2014robust}.

In this work, we propose an online, 
self-supervised\footnote{To be consistent with prior work \cite{bewley2016alextrac}, we use the term self-supervised in that there is no external labeling process for our data. This kind of learning is more properly termed weakly-supervised learning in that a supervised learner algorithm is used with potentially noisy labels automatically derived from the data.} framework for refining deep descriptor models by self-labeling challenging object-triplets in real time. 

We leverage networks pre-trained for image classification, a task for which training data is abundantly available \cite{deng2009imagenet}, to provide an initial descriptor space from which to self-supervise the generation of the labels necessary for training the same models to the challenging task of intra-class object disambiguation in novel domains.
We call our approach DELTA, for Descriptor Enhancement via Labelling Triplets Attentively. 

We exploit the descriptor similarity between detected image patches in consecutive frames, facilitated by their strong visual affinity, to find temporally distant appearances of the same object for positive reinforcement. We further leverage the descriptor space to select difficult negative samples that currently appear to be most similar to the positive example.

We demonstrate the advantages of our method in the context of object tracking-by-detection by evaluating an incrementally refined network through several multiple-object-tracking (MOT) benchmarks. 
Our online approach learns the descriptors, rather than an affinity metric, and experimentally shows improved tracking performance when trained not only by similar objects separated temporally, but also by negative samples near in the descriptor space.
We focus on improving descriptor learning based on visual characteristics alone, observing that our approach can complement methods that consider motion models \cite{milan2014continuous, yoon2018online} or global (rather than incremental) information \cite{yang2012an, ma2018customized}.
Our empirical analysis of a convolutional neural network previously trained for image classification enables a 94\% reduction in model parameters with an improvement in descriptor performance and makes our algorithm tractable for online optimization. \textcolor{black}{While deeper networks have recently demonstrated success in disambiguating inter-class instances \cite{krizhevsky2012imagenet}, this improvement suggests that earlier layers in the classification network may maintain the information required to tell apart intra-class object instances.}

Our method outperforms other approaches that utilize object motion models in terms of multiple object tracking accuracy (MOTA), despite using only incremental visual information \textcolor{black}{and  not assuming motion priors. When compared to other self-supervised refinement methods \cite{ma2018customized}, our approach provides faster adaptation to new data.}

In the following sections, we formulate the self-supervised online descriptor optimization problem, and discuss our triplet cosine loss as well as the procedure for generating training labels. We describe how our algorithm runs in parallel to a traditional frame-to-frame object tracker, incrementally updating the descriptor generation model. Finally, we report results on the challenging 2D-MOT-2015 tracking dataset, and show that we achieve improved MOTA performance despite drastic computational savings and using only visual information.

\section{Data Association and Tracking Problem Overview}

We are interested in the tracking-by-detection problem in dynamic video sequences. There, each observed frame includes bounding-box detections of objects (such as vehicles, pedestrians, cyclists, etc., as illustrated in Fig. \ref{fig_triplet_gen}). Some bounding boxes may also be erroneous detections of the background. We would like to associate each object detection to a previously tracked object, or create new tracks if prior tracked objects are not available. In our framework, we focus on incrementally refining the data-association component of a simple object tracker. Sub-Sections \ref{sec_data_association} and \ref{sec_tracking} formulate the data-association and object tracking problems. Finally, Section \ref{sec_self_super} details our self-supervised approach for incrementally improving data-association performance online. 

\begin{figure*}
    \center
        \includegraphics[scale=0.66]{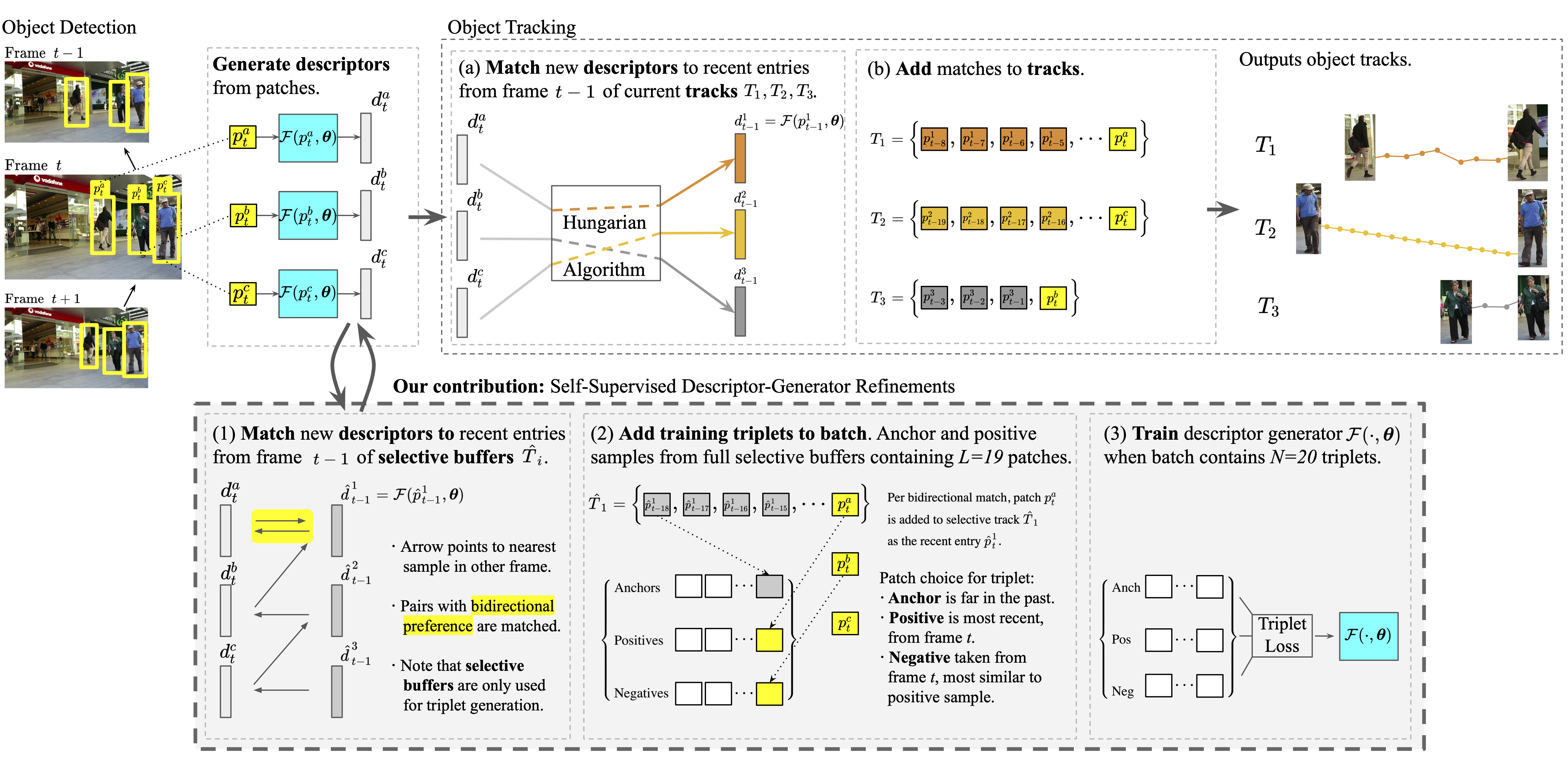}
        \caption{Our proposed system is composed of a descriptor generator, object tracking module, and self-supervised learning pipeline. The descriptor generator $\descgen(\cdot, \boldsymbol{\theta})$ converts object measurements to descriptors, and the two processes (tracking and descriptor refinement) run in parallel and interact only through the learned descriptor generator. The self-supervised method keeps selective buffers $\hat T_j$, which require that bi-directional preference be satisfied, for the purposes of dataset construction. The object tracker is less particular, and matches all new patches to existing tracks $T_j$ to provide the best possible estimates for all detections.}
        
        \label{fig_system_diagram}
\vspace{-0.3in}
\end{figure*}

Our self-supervised descriptor-learning method runs parallel to the object tracker, interacting only via the deep descriptor model. Fig. \ref{fig_system_diagram} illustrates this modular separation, which allows any object tracker using visual descriptors to refine these online using our method. We consider a simple online tracking algorithm to generate the online tracking results, leaving more sophisticated methods for future work.

\begin{figure}[tbp]
\vspace{0.1in}
\centerline{\includegraphics[scale=0.4]{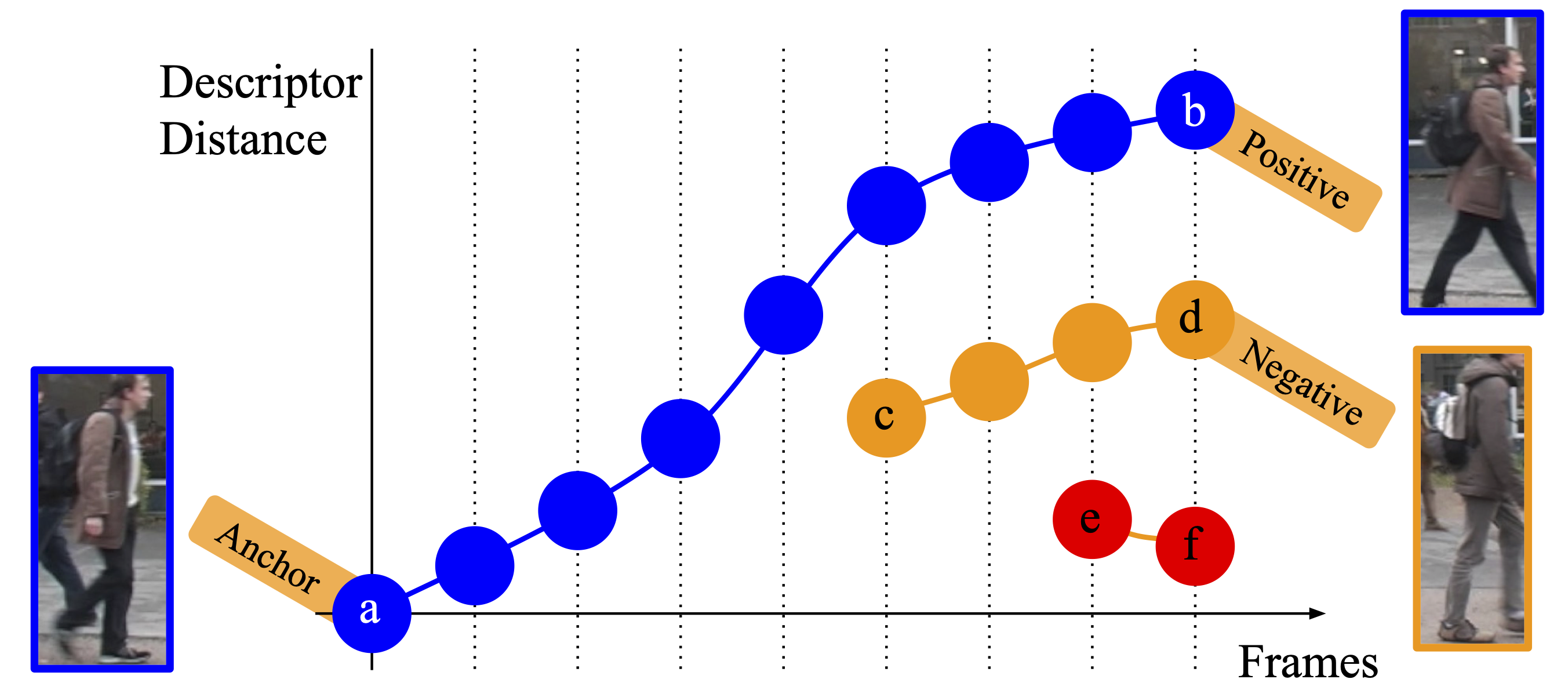}}
\caption{Illustration of triplet selection, where tracked objects are distinguished by color and similarity to anchor ($a$) by vertical distance. To build a challenging triplet with a positive sample at $b$ for the black object, we choose a negative example that is near in the descriptor space ($d$) and an anchor example that is distant temporally ($a$). By choosing $d$ over $f$ for the negative example we generate a more informative nuanced label, as $d$ and $b$ are relatively close in the descriptor space. Object patches extracted from the sequence TUD-Campus in 2D-MOT-2015 dataset \cite{leal2015motchallenge}.}
\label{fig_triplet_gen_ill}
\vspace{-0.25in}
\end{figure}

\subsection{Visual Data Association Problem }

\label{sec_data_association}

Given consecutive image frames of a scene, and bounding-box detections for objects in the frame, we extract the image contents in the boxed sections, i.e., ``patches".
Let $p_t^i$ denote the patch extracted from bounding box $i$ in the frame seen at time $t$. We declare two temporally separated patches $p_{t-1}^i, p_t^j$ as positive match if these are images of the same object instance. To this end, we embed image patches $p_t^i \in \mathbb{R}^{h_t^i \times w_t^i \times 3}$  (of height and width $h_t^i, w_t^i \in \mathbb{N}$)
in a lower dimensional descriptor vector
 $\mathbf{d}_t^i \in \mathbb{R}^n$.

 
Descriptor vectors for patches (resized to $h \times w$) are computed by a mapping $\descgen: \mathbb{R}^{h \times w \times 3} \rightarrow{\mathbb{R}^n}$ parameterized by $\boldsymbol{\theta}$, i.e. patch $p_t^i$ is mapped to descriptor $\desc_t^i = \descgen(p_t^i, \boldsymbol{\theta})$. We define the similarity between object patches to be the distance between the descriptors associated with them, as computed via a distance metric. The similarity between $\desc_{t-1}^i$ and another descriptor $\desc_t^j$ is given by a distance function $\mathcal{D}(\desc_{t-1}^i, \desc_t^j)$, where $\mathcal{D}: \mathbb{R}^n \times \mathbb{R}^n \rightarrow \mathbb{R}$. For patches $p_t^i, p_{t-1}^{i'}$ of the same object, and $p_t^j$ of a different object, we would like the distance function and descriptor model to yield 
\begin{equation}
\mathcal{D}(\desc_{t-1}^i,\desc_t^{i'}) < \mathcal{D}(\desc_{t-1}^i,\desc_t^j).
\label{eq:descriptor_inequality}
\end{equation}
Given an appropriate descriptor space, Equation \ref{eq:descriptor_inequality} intuitively allows for discrimination between similar and dissimilar patch pairs through distance values -- the smaller the distance between patch descriptors is, the more likely they are to correspond to the same object.

The descriptor-generation function $\descgen$ must capture the highly complex mapping between the raw pixel data to the descriptor space, allowing $\mathcal{D}$ to produce meaningful distance values to discriminate between similar and dissimilar object patches. 
It is easy to see that approaches that keep $\descgen$ fixed, and learn an affinity metric $\mathcal{D}$, may fall short when $\descgen$ produces descriptors that cannot be disambiguated under any affinity metric. Therefore, we choose to learn the parameters $\boldsymbol{\theta}$ of the descriptor-generation function.

One method of achieving this complex mapping is by learning the parameters $\boldsymbol{\theta}$ from a labeled training dataset $\mathcal{S}$ where pairs of image patches are labeled as positive (both patches are observations of the same object) or negative pairs (patches of different objects). Given $\mathcal{S}$, a descriptor-generating function $\descgen$ could be trained to minimize the distance $\mathcal{D}(\desc^i,\desc^{i'})$ for similar object patches $p^i, p^{i'}$ by using a fixed or learned distance metric $\mathcal{D}$. However, in practice it is difficult to build such datasets. In this work we therefore label relevant training samples in real time.

\subsection{Object Tracking-By-Detection Problem}
\label{sec_tracking}
For video sequences, where each frame includes noisy bounding box object detections, the object tracking task is to associate each valid bounded image patch, i.e., showing at least part of an object, with a unique identity representing that object. In our framework, we call each unique identifier a ``track", and formulate it as a set of object patches $T_i = \{ \dots, \patch_{t-2}^i, \patch_{t-1}^i\}$ all of the object $i$ before time $t$.

We attempt to match every detected frame-patch to a track using visual similarity alone, without relying on predictive motion models. To this end, the tracking problem reduces to a data association task -- matching observed image patches to existing tracks. 
Under the assumption that the distance $\mathcal{D}$ is smaller for descriptor pairs of the same object than for different objects, we formulate this problem as an optimization, aiming to choose the least-distance assignment between input patches $\patch_t^i$ and tracks $T_j$. 
Let the binary decision variables $x_{i,j}$ take the value $1$ when input patch descriptor $\desc_{t}^i = \descgen( \patch_t^i, \boldsymbol{\theta})$ matched to the most recent entry of track $T_j$, i.e. $\patch^j_{t-1}$, and $0$ otherwise, 
our objective is
\begin{equation}
    \min_\theta \mathcal{D}(\desc_t^i, \descgen(\patch^j_{t-1}, \boldsymbol{\theta})) \cdot x_{i,j},
\label{eq:hungarian}
\end{equation}
under the constraint that as many input patches as possible are matched to tracks. If there exist more detections than tracks, unassigned inputs are each assigned a new tracks. We solve this optimization with the Hungarian algorithm \cite{kuhn1955hungarian}. Our specific implementation details are in Section \ref{sec_experiments}.

\section{Online Self-Supervision}
\label{sec_self_super}

In this section we describe our choice of loss function and distance metric for refining descriptors online, detail our method for choosing difficult positive and negative training samples for online training, and discuss our descriptor-generating model. As detailed in Fig. \ref{fig_system_diagram}, we label patch-triplets in a self-supervised framework and use those to  train our descriptor-generating model. Fig. \ref{fig_triplet_gen_ill} illustrates our use of two sources of information for online dataset construction: time and visual appearance.

\begin{figure}[tbp]
\vspace{-0.05in}
\centerline{\includegraphics[scale=0.17]{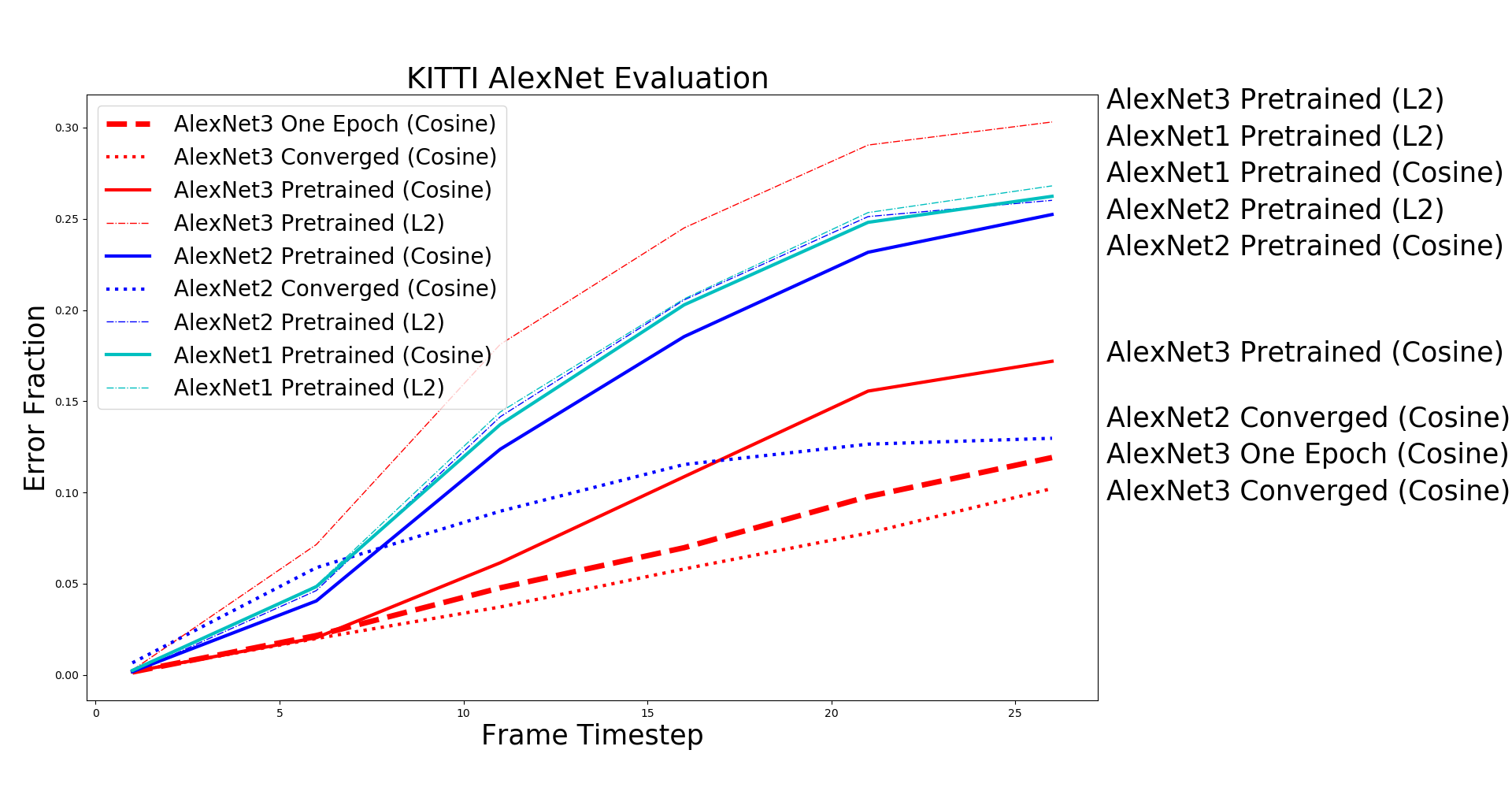}}
\caption{Supervised descriptor evaluation on KITTI \cite{geiger2012are} dataset. We evaluate descriptors extracted from the last max-pooling layer within AlexNet (AlexNet3), and the two fully connected layers that follow it (AlexNet2, AlexNet1). Given two frames at time steps $t$ and $t+\Delta$, for all similar patches $\patch_t^i, \patch_{t+\Delta}^i$ with dissimilar patches  $\patch_{t+\Delta}^j$, we declare an error if $\mathcal{D}_{\cos}(\desc_t^i, \desc_{t+\Delta}^j) \leq \mathcal{D}_{\cos}(\desc_t^i, \desc_{t+\Delta}^i)$. We observe overall worse performance as $\Delta$ grows. ``AlexNet3 One Epoch" was trained on the training samples for one iteration, as opposed to repeating the training to convergence, and shows quick learning of effective descriptors.
Training was performed using ground truth detections in sequences 08-20 divided to 1740 batches of 20 triplets where the positive and anchor samples were 15 and 20 frames apart. The evaluation above was done on the remaining sequences in 80460 comparisons. Our learning rates were $10^{-4}$ for ``AlexNet3 One Epoch" and $10^{-10}$ for the converged models.}
\label{fig_delta_ts}
\vspace{-0.25in}
\end{figure}
\subsection{Descriptor Refinement with Cosine Triplet Loss}
\label{sec_triplet}

In order to learn a complex mapping $\descgen$ between object patch pixels to descriptors, a Siamese \cite{bromley1994signature} set of neural networks is often used \cite{yoon2018online}. Two or more identical descriptor-generators learn to produce different outputs by training on image-patches labeled as similar or dissimilar. Loss functions, such as contrastive loss \cite{chopra2005learning}  for training pairs or triplet-loss \cite{weinberger2009distance} for training triplets, combine descriptors to a single loss value. Given a descriptor distance metric $\mathcal{D}$, both loss functions aim to minimize the distance 
between similar descriptors and to maximize the distance between dissimilar pairs. We choose to use the triplet loss function since it has an inherent balance of positive and negative samples, and found it easier in practice to optimize than a contrastive loss. Additionally, the cosine triplet loss has proven useful in supervised learning contexts \cite{kavitha2019evaluation}.

Given $\mathbf{d_\text{anchor}}, \mathbf{d_\text{positive}}$ for descriptors of the same object and $\mathbf{d_\text{negative}}$ for a descriptor of a different object, the triplet loss is
\begin{multline}
\vspace{-0.2in}
L_{\text{triplet}}(\desc_{\text{anchor}},\desc_{\text{positive}},\desc_{\text{negative}}) =\\ \max \{0, \mathcal{D}(\mathbf{d_\text{anchor}}, \mathbf{d_\text{positive}})  + m - \mathcal{D}( \mathbf{d_\text{anchor}}, \mathbf{d_\text{negative}} ) \}
\label{eq:triplet_loss}
\end{multline}
where $m \in \mathbb{R}$ is a margin parameter marking sufficient dissimilarity between negative pairs.
Although the distance metric $\mathcal{D}$ could be learned \cite{zagoruyko2015learning}, we elect to use the fixed cosine distance metric (Equation \ref{eq:cos}), given that our work focuses on learning descriptors. 
\begin{equation}
\begin{aligned}
\mathcal{D}_{\cos}(\desc^i, \desc^j)= 1 - \dfrac{\desc^i \cdot \desc^j}{\Vert \desc^i \Vert \Vert \desc^j \Vert} .
\end{aligned}
\label{eq:cos}
\end{equation}
As depicted in Fig. \ref{fig_delta_ts}, we experimentally verified the results from \cite{kavitha2019evaluation}, showing that the cosine distance $\mathcal{D}_{\cos}$ metric outperformed Euclidean distance in the supervised task of frame-to-frame data association.

\subsection{Positive Sample Collection}
\label{sec_positive}

To construct our triplets, we seek to find pairs of image patches of the same object. We seek positive pairs that are expected to be difficult to classify as similar, to provide a good training signal to our model, i.e., we would like to find $\patch_{\text{anchor}}$ and $\patch_{\text{positive}}$ such that the initial distance  $\mathcal{D}(\desc_{\text{anchor}}, \desc_{\text{positive}})$ is large.

We collect these difficult pairs by choosing temporally distant image patches.
We harness the high frame-rate of video sequences to track objects whose appearance does not vary dramatically between consecutive frames -- a property keeping their descriptors close in the initial embedding space. As shown in Fig. \ref{fig_system_diagram}-(1), we create the bounded-length ``selective" patch buffers $\hat T_j$ independently from the object tracker by performing bidirectional matching between descriptors $\desc_t^i$ of new object detections and the descriptors for the recent entries of selective buffers $\hat T_j$, namely $\hat \desc_{t-1}^{j}=\descgen(\hat p^{j}_{t-1} ,\boldsymbol{\theta})$. Among $Z$ frame detection patches, we consider associating a patch $\patch_t^i$ to any of the $M$ selective buffers $\hat T_j$, making it the buffer's most recent entry $\hat \patch^j_t$, if
\begin{equation}
\textbf{AND}
    \begin{cases}
    \argmin \limits_{m \in \{1,\dots M\}}  \mathcal{D}(\desc_t^i, \hat \desc^{m}_{t-1}) = j \\
    \argmin \limits_{z \in \{1,\dots Z\}}  \mathcal{D}(\hat \desc^{j}_{t-1}, \desc_t^z) = i \\
    \end{cases} \;\;\;\;\; .
\end{equation}
Unlike the Hungarian algorithm, which matches as many detections to tracks as possible when solving Equation \ref{eq:hungarian}, this criterion is designed to be more selective. Buffers not extended are deleted and unassigned detection patches join new empty buffers. When a buffer $\hat T_j$ contains $L$ patches thought to be of consecutive observations of the same object, we choose $\patch_{\text{anchor}},\; \patch_{\text{positive}}$ to be the temporally distant pair $\hat \patch^j_{t-L+1}, \;\hat\patch^j_{t}$. Full buffers accept new entries and discard old ones to maintain a size of $L$.

We use these selective buffers, which often include easily tracked objects, to improve the descriptors used to match more challenging objects.

\subsection{Negative Sample Collection}
\label{sec_negative}

To increase the utility of our limited training samples, we find negative samples that are difficult to disambiguate from the positive sample using the descriptor distance metric $\mathcal{D}$. We want a $\patch_{\text{negative}}$ such that $\mathcal{D}(\mathbf{d}_{\text{positive}}, \mathbf{d}_{\text{negative}})$ is currently small, so that we can improve the descriptors to create more discriminative distance between the positive and negative classes. As shown in Fig. \ref{fig_system_diagram}-(2) we consider patches detected in frame $t$, the frame from which the positive sample is drawn from, as possible negative sample candidates. We choose the detection patch $\patch_t^i$ whose descriptor $\descgen(\patch_t^i, \boldsymbol{\theta})$ is most similar to the positive patch descriptor $\desc_{\text{positive}}$,

\begin{equation}
    \patch_{\text{negative}} = \argmin \limits_{\patch_t^i \in \{\text{frame } t\} \textcolor{black}{\setminus \{ \patch_{\text{positive}}\}}} \mathcal{D}(\desc_{\text{positive}}, \descgen(\patch_t^i, \boldsymbol{\theta})).
\end{equation}
\subsection{Online Optimization}
As shown in Fig. \ref{fig_system_diagram}, we generate training triplets in a process parallel to online tracking. After accumulating $\varSizeBatch$ training examples, we train the descriptor model $\descgen(\cdot, \boldsymbol{\theta})$ for a single epoch, optimizing Equation \ref{eq:train} by substituting in Equations \ref{eq:triplet_loss}, \ref{eq:cos}, i.e.,
\begin{equation}
    \label{eq:train}
    \begin{split}
    \argmin \limits_{\boldsymbol{\theta}}\dfrac{1}{N} \sum_i^\varSizeBatch L_{triplet}(&\descgen(p_{\text{anchor}}^i,\boldsymbol{\theta}),\\
    &\descgen(p_{\text{positive}}^i, \, \boldsymbol{\theta}), \descgen(p_{\text{negative}}^i, \boldsymbol{\theta})).
    \end{split}
\end{equation}
We discard each batch after optimization. Our parameter choices are detailed in Section \ref{sec_implementation_details}.

\section{Adaptive Descriptor Generators}
\label{sec_descgens}

CNNs with demonstrated performance in image classification competitions such as \cite{deng2009imagenet} are commonly used for image patch matching applications \cite{kavitha2019evaluation, yuan2019robust}. Although these networks \cite{krizhevsky2012imagenet, simonyan2014very} are pre-trained for classification, rather than instance-level disambiguation, previous works have shown that the utility of such networks to generate reasonable descriptor spaces \cite{bewley2016alextrac}. We propose leveraging the initial suitability of these networks for refining descriptor generation online, and therefore seek network architectures suitable for real-time use and fast adaptation for online refinement. Similar to previous work, we consider the activation values within one of the network's hidden layers as a descriptor for the image patch input to the network, and utilise a distance metric $\mathcal{D}$ to determine similarity between descriptors.

Our experiments in Section \ref{sec_experiments} use the output of the last max-pooling layer within AlexNet \cite{krizhevsky2012imagenet} as our descriptor-generator, and call the subset of AlexNet up to this layer AlexNet3, given that it is missing three fully-connected layers. This network includes only 6\% of the parameters originally used in AlexNet, 
and compared to VGG-16 \cite{simonyan2014very} as it is used in \cite{bewley2016alextrac} (we name it ``VGG-16-2'', since it misses two layers). AlexNet3 has 3.2\% of the parameters of ``VGG-16-2''.
We include an evaluation of VGG-16-2 as a descriptor generator in Section \ref{sec_experiments} as well.

Fig. \ref{fig_delta_ts} details our supervised experimental evaluation of AlexNet in the context of visual data-association across long time steps in videos, and shows that it initially performs about 35\% better relatively to other evaluated hidden layers, when pretrained on Imagenet \cite{deng2009imagenet} only -- without online refinements. These results suggest that earlier layers in this classification network may maintain more detail required to disambiguate intra-class object instances. Additionally, the figure demonstrates AlexNet3 pretrained on Imagenet and tested on KITTI can reduce its error rates by additional 35\% after a single supervised training epoch on ground truth bounding boxes from a reserved subset of KITTI \cite{geiger2012are}.
AlexNet3's ability to adapt to new data quickly makes it a favorable choice for a self-supervised descriptor-generator. 

\section{Experiments}
\label{sec_experiments}

We evaluated the performance of our proposed method for online descriptor enhancement via self-labelling triplets attentively (DELTA) in the context of tracking-by-detection, as shown in Table \ref{table_results}. We evaluate tracking performance in the dynamic video sequences of the popular 2D-MOT-2015 dataset \cite{leal2015motchallenge}, as it provides the temporal structure necessary for constructing dataset triplets online. In the following subsections we discuss the evaluation dataset, the implementation details of our experimental framework, and our performance as compared to baseline methods.

\begin{table*}[]
\vspace{0.1in}
\label{table_results}
\begin{center}
\begin{tabular}{|l|l|l|l|l|l|l|}
\hline
\textbf{Set} & Method & MOTA$\uparrow$ & MOTP $\uparrow$   &IDs$\downarrow$  &  FP$\downarrow$ & FN$\downarrow$   \\ \hline

\hline

\multicolumn{1}{|l|}{\textit{Train} set}  &DELTA + AlexNet3 (\textbf{ours})    & 27.8\%  & 71.2\% & 480 & 6097 & 22228  \\ \cline{2-7} 
\multicolumn{1}{|l|}{results AlexNet3}  &DELTA Easy-Positives    & 20.5\%  & 71.9\% & 843 &  10349	& 20533 \\ \cline{2-7}
\multicolumn{1}{|l|}{}  &DELTA Random-Negatives   & 26.9\%  & 72.1\% & 516 &  6507 & 22146 \\ \cline{2-7}
\multicolumn{1}{|l|}{}  &AlexNet3 Pretrained    & 20.5\%  & 71.9\% & 843 &  10349	& 20533   \\ \cline{1-7}
\multicolumn{1}{|l|}{\textit{Train set}}  &VGG-16-2 Pretrained   & 19.0\%  & 71.9\% &  1032& 10974  & 20337  \\ \cline{2-7}
\cline{2-7}
\multicolumn{1}{|l|}{results VGG-16-2}  &DELTA + VGG-16-2   & 26.1\%  & 72.2\% & 551 & 6932 & 22008 \\ \cline{2-7}
\hline
\hline

\hline

\multicolumn{1}{|l|}{\textit{Test} set results for} & DELTA + AlexNet3 (\textbf{ours})  &\textbf{21.25\%}  & 70.95\%   & \textbf{1231}  &  \textbf{8597} & \textbf{38557}  \\ \cline{2-7}
\multicolumn{1}{|l|}{strictly visual methods} & ALEX-TRAC \cite{bewley2016alextrac}  & 17.0\%  & \textbf{71.2\%}   & 1859 &   9233 & 39933 \\ 
\hline
\hline

\hline
\multicolumn{1}{|l|}{Additional \textit{Test} set} & TC\_SIAMESE$$ \cite{yoon2018online}  & 20.2\%  & 71.1\%   & 294  &  6127 & 42596  \\ \cline{2-7}
\multicolumn{1}{|l|}{results} & TBD$$ \cite{geiger20133d}  & 15.9\%  & 70.9\%   & 1939  &  14943 &	34777  \\ \cline{2-7}
\multicolumn{1}{|l|}{} & TC\_ODAL$$ \cite{bae2014robust}  & 15.1\%  & 70.5\%   &637  &   12970 & 38538	   \\ \cline{2-7}
\multicolumn{1}{|l|}{  } & LDCT$$ \cite{solera2015learning}  & 4.7\%  & 71.7\%   & 12348 &   14066 & 32156  \\ 
\hline


\end{tabular}
\end{center}
\vspace{-0.1in}
\caption{2D-MOT-2015 Results. When compared to other trackers emphasizing data-association, and in particular to ALEX\_TRAC as it is concerned with strictly visual data association, we achieve better multiple object tracking accuracy (MOTA), a metric which comprehensively combines several statistics as shown in Equation \ref{eq:mota}. We report the overall number of false-positives (FP), false-negatives (FN), and identity-switches (IDs), and list the multiple object tracking precision (MOTP $\uparrow$) metric. We write $\uparrow$ next to metrics where larger value is better, and $\downarrow$ where smaller value is preferred.}
\label{table_results}
\vspace{-0.26in}
\end{table*}

\subsection{Implementation Details}
\label{sec_implementation_details}

In our evaluation we used the Hungarian algorithm with the cosine distance $\mathcal{D}_{\cos}$ to match detected patches in frames to existing tracks. 
We chose AlexNet3 (see Section \ref{sec_descgens}) as our descriptor-generator $\descgen$, and refined the model online with DELTA. We implemented our approach in Python, with the network implementations provided by PyTorch 
and the Hungarian algorithm by SciPy
\cite{virtanen2020SciPy}
. Extracted patches were resized to  $227 \times 227$ pixels and normalized to the PyTorch input standards for ImageNet models. 

We aggregated batches of 20 triplets for each training cycle, where we used a learning rate of $3.28 \times 10^{-5}$ and margin $m$ of $0.3$. Additional parameters were obtained via a black-box optimization \cite{knyshK2016blackbox} over the \textit{Train} set assuming AlexNet3 as the descriptor-generator. Matches with cost larger than a threshold $0.59$ were discarded. Tracks $T_i$ not extended for more than one frame were de-registered and not extended further. If upon de-registration a track's length was less than 12 frames, it was erased and not used in evaluation. The temporal distance $L$ between anchor and positive samples was 19 frames. All results reported for DELTA, for both on the \textit{Train} and \textit{Test} datasets and both AlexNet3 and VGG-16-2, use the same parameter values.

\subsection{Evaluation Datasets}
We tested DELTA using the 2D-MOT-2015 multiple-object-tracking (MOT) \cite{leal2015motchallenge} and the MOT-16 \cite{milan2016mot16} benchmarks. These datasets provide a number of video sequences along with bounding-box detections of the objects their frames as generated by the object detector \cite{dollar2014fast}. These challenging benchmarks include videos where the camera is static and dynamic, videos where objects are often occluded, and generally imperfect bounding box detections that include partial detections of objects (i.e., only a leg of a person) or of background. \textcolor{black}{Our results from these diverse videos shed light on the ability of our self-supervised method to improve tracking performance in different settings.}

We quantify the performance in the tracking task, as it is formulated in Section \ref{sec_tracking}, using the widely accepted CLEAR-MOT metrics \cite{bernardin2008evaluating}. There, the multiple object tracking accuracy (MOTA $\uparrow$) metric quantifies tracking performance comprehensively with the relationship
\begin{equation}
    \text{MOTA} = 1-\dfrac{\sum_t ( \mathrm{FP}_t + \mathrm{FN}_t + \mathrm{IDs}_t )}{\sum_t\mathrm{GT}_t}.
\label{eq:mota}
\end{equation}
At frame t, $\mathrm{FP}_t$ is the number of false positives (patches not showing an object) included in tracks, $\mathrm{FN}_t$ is the number of false negatives (correct detections not included in tracks), $\mathrm{IDs}_t$ is the number of times a true object is assigned a different identity, and $\mathrm{GT}_t$ is the number of ground-truth objects.
The multiple object tracking precision (MOTP $\uparrow$) metric is also included, and represents the spatial misalignment of reported tracks and true tracks.

The dataset was divided to \textit{Train} and \textit{Test} sets. We ran our proposed incremental approach on each sequence in the dataset, resetting the network to its pre-trained state after the sequence terminated and before the next was evaluated. The \textit{Train} set can be evaluated locally and was used for choosing values via black-box optimization for the fixed tracker parameters specified in subsection \ref{sec_implementation_details}. We did not use ground-truth bounding boxes for any part of our method.

\subsection{Evaluation Methods}
We primarily compare DELTA to other works where visual data association is evaluated by tracking performance. Most similar to our work is the self-supervised tracker ALEX\_TRAC \cite{bewley2016alextrac}, where only visual information is used to train an affinity metric online. Similar to our method, they also generate descriptors with a subset of a CNN pretrained on Imagenet \cite{deng2009imagenet} (i.e., VGG-16-2) and use the Hungarian algorithm as the main data association module for object tracking. The most prominent difference between the two methods is DELTA learns to refine our descriptor-generating network online while ALEX\_TRAC learns an affinity metric. 

We also include results from relevant trackers that make use of motion models in addition to visual matching. TC\_ODAL \cite{bae2014robust} trains a descriptor-generator offline and refines it online, and TC\_SIAMESE \cite{yoon2018online} learns a distance metric offline. Both require video datasets with relevant objects and ground-truth detections for training. The self-supervised method LDCT \cite{solera2015learning} learns tracker parameters in a Latent Structural SVM framework.
We also include the general tracking method TBD \cite{geiger20133d}, which utilizes the Hungarian algorithm.

To assess the efficacy of our proposed method for generating difficult positive and negative examples, we evaluated DELTA with one frame-step between anchor and positive samples (denoted ``DELTA Easy-Positives") --  making the collected positive samples more visually similar to the anchors. We tested a version of our algorithm where negative samples are randomly chosen ``DELTA Random-Negatives''. We applied DELTA to refine a different descriptor generator VGG-16-2 (``DELTA + VGG-16-2"), and included the performance achieved with Imagenet pre-training alone, without online optimization, at ``AlexNet3 Pretrained'' and ``VGG-16-2 Pretrained", for reference. 

\textcolor{black}{Finally, we add a comparison to HCC \cite{ma2018customized}, a pedestrian tracker pretrained extensively on domain-specific data where local tracks are globally optimized for consistency after processing videos. Similarly to our method, HCC collects self-labelled observed patches from the videos for training a data association network. Given that HCC harnesses both spatial and visual global information, we do not attempt to outperform their MOTA score, but aim to compare the \textit{gain} in the MOTA scores facilitated by our self-supervised refinement module to theirs (column $\Delta$MOTA in Table \ref{table_results_mot_16}). We note that we only use information available during tracking-time, and use a network 50\% smaller than HCC's that was pretrained with domain-independent data. As we discuss in Fig. \ref{fig_delta_ts}, the smaller network facilitates quick and effective learning.}




\begin{table*}[]
\vspace{0.1in}
\begin{center}
\begin{tabular}{|l|l|l|l|l|l|l|}
\hline
\textbf{Sequence} & Method & $\Delta$MOTA$\uparrow$ & MOTA $\uparrow$   &IDs$\downarrow$  &  FP$\downarrow$ & FN$\downarrow$   \\ \hline

\hline

\multicolumn{1}{|l|}{MOT16-11}  &DELTA + AlexNet3 (\textbf{ours})    & \textbf{5.1\%}  & 45.5\% & 61 & 730 & 4205  \\ \cline{2-7} 
\multicolumn{1}{|l|}{}  &HCC    & 0.9\%  & 55.1\% & 8 &  352	& 3762 \\ \cline{1-7}
\hline
\hline

\hline
\multicolumn{1}{|l|}{MOT16 \textit{Train} Set}  &DELTA + AlexNet3 (\textbf{ours})   &\textbf{ 6.2\% }& 24.8\% & 753 & 12630 & 69611 \\ \cline{2-7}\hline 
\hline

\hline
\multicolumn{1}{|l|}{MOT16 \textit{Test} Set}  &DELTA + AlexNet3 (\textbf{ours})   & & 32.6\% & 1545 & 15305 & 106060 \\ \cline{2-7}
\multicolumn{1}{|l|}{}  &HCC   &   & 49.3\% & 391 & 5333 & 86795 \\ \cline{2-7}
\hline

\end{tabular}
\end{center}
\vspace{-0.1in}
\caption{\textcolor{black}{MOT-16 results. We present MOTA gains achieved by refining AlexNet3 online using our method. Additionally, we show our MOTA improvement is significant by comparing it to results by HCC \cite{ma2018customized}, a pedestrian tracker pretrained on extensive domain-specific data that uses a self-supervised data collection module to match local tracks into consistent global tracks after observing the video. Given that our objective is aimed at improving data association online and during tracking-time, we do not attempt to outperform HCC's tracking results, but rather argue that DELTA outperfoms HCC in terms of performance \textit{gain}.
The $\Delta$MOTA column shows the MOTA score gained by using the self-supervised refinement module in DELTA (ours) and HCC, compared to performing tracking without refinement. Only available for the training set. The other columns are similar to Table \ref{table_results}}.}
\label{table_results_mot_16}
\vspace{-0.26in}
\end{table*}

\subsection{Experimental Results}

Among all evaluated methods in 2D-MOT-2015, DELTA with AlexNet3 achieves the highest \textit{Test} set MOTA score of $21.25\%$, despite not making use of motion models or any spatial information. 
Shown in Table \ref{table_results}, we outperform ALEX\_TRAC despite using a substantially smaller network, \textcolor{black}{showing that a relatively shallow network, which is necessary for online optimization, is also sufficient for discrimination between visual object observations when trained with effective data collected online with our approach.}

We show that finding hard positive and hard negative examples is useful for optimizing descriptor performance.
\textcolor{black}{As shown in Table \ref{table_results_mot_16}, the naive sample selection use in HCC for refining their global track fusion module improved their MOTA score by $1\%$ on the MOT-16-11 sequence. On average, DELTA yields a gain of $6.2\%$ on the MOT-16 training videos, suggesting that purposeful difficult sample selection for online refinement of the local tracking module could benefit the global performance.} Additionally, Table \ref{table_results} shows that choosing positive samples from shorter selective buffers (``DELTA Easy-Positives")
had practically no effect on the network, achieving a similar score to that achieved without online learning (``AlexNet3 Pretrained"). In addition, allowing negative samples to be randomly chosen (``DELTA Random-Negatives") also results in a degradation in performance, confirming the utility of selecting negative examples that are near in the descriptor space to the anchor. 

We also demonstrate the applicability of our online training procedure and network selection; we show that DELTA improves MOTA scores for both ``DELTA + AlexNet3 (\textbf{ours})" and ``DELTA + VGG-16-2" from their respective pre-trained performances (``AlexNet3 Pretrained", ``VGG-16-2 Pretrained"). However, ``DELTA + VGG-16-2" averages a processing rate of 0.1 frames per second, while our analysis in Section \ref{sec_descgens} enables our approach to process 5.5 frames per second on average, further emphasizing the effectiveness of DELTA in refining small descriptor generators.

We note that the MOTP score, which quantifies spatial deviation of object tracks from their true positions, of all evaluated methods is relatively similar, within a margin of $0.7\%$. The consistency may arise because all methods make use of the same bounding box detections provided by the dataset.
Additionally, Table \ref{table_results} shows that methods relying on motion models can achieve better IDs scores -- an expected result given that motion predictions are more robust to occlusions -- shedding light on the potential our method has for solving the visual data association task in trackers leveraging motion models in future work. 

\section{Conclusion}
\label{sec_conclusion}
We have presented a novel method for incrementally refining object-descriptor-generators online without the need of labeled data or prior domain knowledge. We have shown that refining descriptors online can help improve visual data association performance, and demonstrated that our approach can be applied to object tracking. By achieving object tracking accuracy better than existing methods, we believe that our self-supervised method can be used to solve the problem of visual data association in object trackers that make use of sophisticated motion models.

\printbibliography
\end{document}